\documentclass{article}
\usepackage{spconf,amsmath,epsfig}

\let\OLDthebibliography\thebibliography
\renewcommand\thebibliography[1]{
  \OLDthebibliography{#1}
  \setlength{\parskip}{0pt}
  \setlength{\itemsep}{0pt plus 0.3ex}
}

\usepackage{pifont}

\newcommand{\cmark}{\ding{51}}%
\usepackage{graphicx}
\usepackage{amsmath}
\usepackage{amssymb}
\usepackage{booktabs}

\usepackage{amsmath,amsfonts}
\usepackage{algorithmic}
\usepackage{algorithm}
\usepackage{array}
\usepackage{textcomp}
\usepackage{stfloats}
\usepackage{url}
\usepackage{verbatim}
\usepackage{graphicx}
\usepackage{cite}
\usepackage{xspace}
\makeatletter
\DeclareRobustCommand\onedot{\futurelet\@let@token\@onedot}
\def\@onedot{\ifx\@let@token.\else.\null\fi\xspace}

\makeatother

\usepackage{amsmath}
\usepackage{amssymb}
\usepackage[table]{xcolor}
\usepackage[pagebackref=true,breaklinks=true,colorlinks=true,bookmarks=false,linkcolor={red!50!black},urlcolor={darkgray!80!black},citecolor={green!50!black}]{hyperref}
\usepackage[group-separator={,}]{siunitx}
\usepackage{multirow}
\usepackage{tabularx}
\usepackage{booktabs} 
\usepackage{makecell}
\usepackage{graphbox}
\usepackage{footmisc}
\usepackage{bm}
\usepackage{caption}
\usepackage{arydshln}
\usepackage{dashrule}
\usepackage{subcaption}
\usepackage{enumitem}
\setlist[itemize]{noitemsep,nolistsep}

\usepackage[capitalize]{cleveref}
\crefname{section}{Sec.}{Secs.}
\Crefname{section}{Section}{Sections}
\Crefname{table}{Table}{Tables}
\crefname{table}{Tab.}{Tabs.}
\Crefname{figure}{Figure}{Figures}
\crefname{figure}{Fig.}{Figs.}
\Crefname{equation}{Equation}{Equations}
\crefname{equation}{Eq.}{Eqs.}

\usepackage{graphicx}
\usepackage{amsmath}
\usepackage{amssymb}
\usepackage{booktabs}

\usepackage{graphicx}
\usepackage{makecell}
\usepackage{comment}
\usepackage{amsmath,amssymb} 
\usepackage{color}
\usepackage{booktabs}
\usepackage{arydshln}
\usepackage{multirow}
\usepackage{hyperref}

\usepackage[]{lineno}

\renewcommand{\paragraph}[1]{ 
 \noindent\textbf{#1}~}

\newcommand{\blue}[1]{\textbf{\textcolor{mblue}{#1}}}

\newcommand{\bred}[1]{\textbf{\textcolor{red}{#1}}}
\newcommand{\green}[1]{\textcolor{mgreen}{#1}}

\newcommand{\gray}[1]{\textcolor{mgray}{#1}}

\colorlet{lightcyan}{cyan!10}
\colorlet{lightpink}{pink!20}
\colorlet{lightgray}{gray!10}

\definecolor{mgray}{gray}{0.35}
\definecolor{mred}{RGB}{238, 34, 12}
\definecolor{mgreen}{RGB}{1, 127, 0}
\definecolor{mblue}{RGB}{0, 77, 128}
\definecolor{orange}{RGB}{240, 120,0}

\usepackage[table]{xcolor}
\usepackage[pagebackref=true,breaklinks=true,colorlinks=true,bookmarks=false]{hyperref}

\usepackage[capitalize]{cleveref}
\crefname{section}{Sec.}{Secs.}
\Crefname{section}{Section}{Sections}
\Crefname{table}{Table}{Tables}
\crefname{table}{Tab.}{Tabs.}

\begin{document}\sloppy

\def\x{{\mathbf x}}
\def\L{{\cal L}}

\title{Exploring Opinion-Unaware Video Quality Assessment \\with Semantic Affinity Criterion}
%
\names{Haoning Wu$^1$\qquad Liang Liao$^1$\qquad Jingwen Hou$^1$\qquad Chaofeng Chen$^1$\qquad Erli Zhang$^1$}{Annan Wang$^1$\qquad Wenxiu Sun$^2$\qquad Qiong Yan$^2$\qquad Weisi Lin$^1$}
\address{}

\maketitle

\begin{abstract}

{
Recent learning-based video quality assessment (VQA) algorithms are expensive to implement due to the cost of data collection of human quality opinions, and are less robust across various scenarios due to the biases of these opinions. This motivates our exploration on opinion-unaware (a.k.a zero-shot) VQA approaches.
Existing approaches only considers low-level naturalness in spatial or temporal domain, without considering impacts from high-level semantics. In this work, we introduce an explicit semantic affinity index for opinion-unaware VQA using text-prompts in the contrastive language-image pre-training (CLIP) model. We also aggregate it with different traditional low-level naturalness indexes through gaussian normalization and sigmoid rescaling strategies. Composed of aggregated semantic and technical metrics, the proposed \underline{B}lind \underline{U}nified \underline{O}pinion-U\underline{na}ware \underline{V}ideo Quality \underline{I}ndex via \underline{S}emantic and \underline{T}echnical Metric \underline{A}ggregation (\textbf{BUONA-VISTA}) outperforms existing opinion-unaware VQA methods by at least \textbf{20\%} improvements, and is more robust than opinion-aware approaches.}
\end{abstract}
%
%
\section{Introduction}
\label{sec:intro}

With the rapid growth in the number of online videos, objective Video quality assessment (VQA) is gaining a great deal of interest from researchers. In recent years, although opinion-aware VQA approaches~\cite{tlvqm,fastvqa,vsfa,cnntlvqm} have been extensively explored, they rely on large amounts of training data with expensive human subjective scores~\cite{pvq,cvd,kv1k,vqc} and are typically not easily adaptable to new datasets. How to alleviate the burden of costly training data and build a robust VQA capable of evaluating any given video is the issue that urgent to study.


\begin{figure}
    \centering
\includegraphics[width=0.93\linewidth]{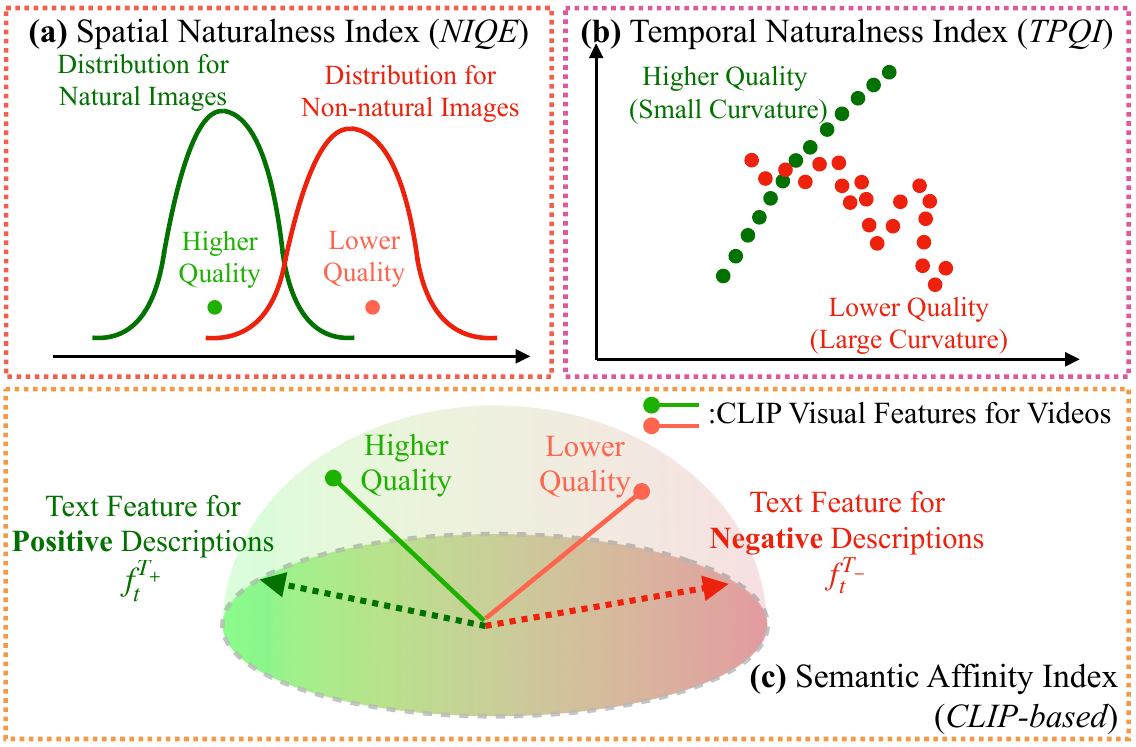}
    \vspace{-9pt}
    \caption{Visualization on  criteria of the three independent metrics in \textbf{BUONA-VISTA}. The pipeline in shown in Fig.~\ref{fig:1}.}
    \label{fig:criterion}
    \vspace{-18pt}
\end{figure}

\let\thefootnote\relax\footnotetext{$^1$Nanyang Technological University; $^2$Sensetime Research.}
\let\thefootnote\relax\footnotetext{$^*$Code available at {{\textit{https://github.com/QualityAssessment/BVQI.}}}}

In recent years, few studies have been conducted in \textbf{opinion-unaware VQA}~\cite{niqe,ilniqe,brisque,tpqi}, which typically rely on empirical criteria for VQA instead of regression on opinion data. For example, NIQE~\cite{niqe} measures \textit{spatial} naturalness of images by comparing them with distributions of pristine natural contents (Fig.~\ref{fig:criterion}(a)). TPQI~\cite{tpqi}, inspired by knowledge on human visual system, measures the \textit{temporal} naturalness of videos through the inter-frame curvature on perceptual domains~\cite{primaryv1,lgn}. These opinion-unaware VQA methods are based on \textbf{\textit{low-level}} criteria, ignoring the human perceptions of video semantics. Moreover, many existing studies have noticed that natural authentic distortions~\cite{spaq,paq2piq,ytugc} or aesthetic-related issues~\cite{vsfa,dover} commonly occur on in-the-wild videos and impact human quality perception. These issues are hardly captured with these low-level criteria, but could be better extracted with semantic-aware deep neural features\cite{cnntlvqm,fastervqa,videval,dbcnn,bvqa2021,mdtvsfa}.


In this paper, we propose a semantic-aware criterion to tackle with these high-level quality issues in an unsupervised manner. With the Contrastive Language-image Pre-training (CLIP)~\cite{clip}, we are able to calculate the affinity between visual features and any given texts. Based on CLIP, we measure whether visual features of a video are more similar to text features for positive (\textit{e.g. high quality}), or negative (\textit{e.g. low quality}) text descriptions (Fig.~\ref{fig:criterion}(c)), which acts as semantic-aware quality criterion mostly focusing on aesthetic-related quality issues and high-level human quality perception. With the new criteria, we design the Semantic Affinity Index as a semantic-aware zero-shot VQA index to assess authentic distortions and aesthetic issues.  
Furthermore, we design the gaussian normalization followed by sigmoid rescaling~\cite{vqeg}, to aggregate the Semantic Affinity Index with low-level spatial and temporal naturalness metrics, composing into the overall Blind Unified Opinion-Unaware Video Quality Index via Semantic and Technical Metric Aggregation (\textbf{BUONA-VISTA}).




In general, our contributions are three-fold:
\begin{enumerate} [topsep=0pt,itemsep=0pt,parsep=0pt]
\renewcommand{\labelenumi}{\theenumi)}
    \item We introduce a novel text-prompted Semantic Affinity Index for opinion-unaware VQA. It incorporates acronym-differential affinity and multi-prompt aggregation to accurately match human quality perception.
    \item We introduce gaussian normalization and sigmoid rescaling strategies to align and aggregate the Semantic Affinity Index with low-level spatial and temporal technical metrics into the \textbf{BUONA-VISTA} quality index.
    \item The BUONA-VISTA significantly outperforms existing zero-shot VQA indexes ($>$20\%), and proves superior robustness against opinion-aware VQA methods.
\end{enumerate}

\section{The Proposed Method}

\begin{figure*}
    \centering
    \includegraphics[width=0.97\textwidth]{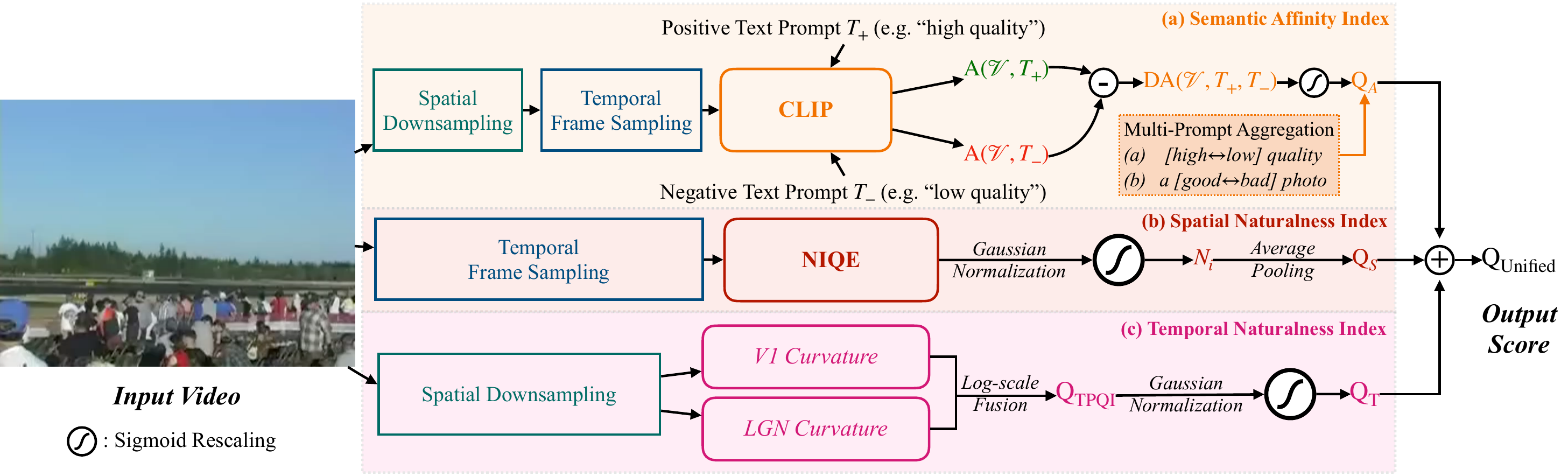} 
    \vspace{-10bpt}
    \caption{The overall pipeline of BUONA-VISTA, including (a) Semantic Affinity Index, (b) Spatial Naturalness index, and (c) Temporal Naturalness Index. The three indexes are remapped and aggregated to the final BUONA-VISTA index.}
    \label{fig:1}
    \vspace{-15pt}
\end{figure*}

In this section, we introduce the three metrics with different criteria that make up the proposed video quality index, including the CLIP-based Semantic Affinity Index ($\mathrm{Q}_A$, Sec.~\ref{sec:qa}), and two technical naturalness metrics: the Spatial Naturalness Index ($\mathrm{Q}_S$, Sec.~\ref{sec:qs}), and the Temporal Naturalness Index ($\mathrm{Q}_T$, Sec.~\ref{sec:qt}). The three indexes are sigmoid-rescaled and aggregated into the proposed \textbf{BUONA-VISTA} quality index. The overall pipeline of the index is illustrated in Fig.~\ref{fig:1}.

\subsection{The Semantic Affinity Index ($\mathrm{Q}_A$)}
\label{sec:qa}
To extract semantic-related quality issues in VQA, we utilize CLIP to extract the \textbf{Semantic Affinity Index} ($\mathrm{Q}_A$) as follows.

\paragraph{Aesthetic-specific Data Preparation.}
As the semantic branch of BUONA-VISTA aim at authentic distortions and aesthetic issues which are usually insensitive to resolutions or frame rates, we follow the data preparation in DOVER~\cite{dover} to perform \textbf{\textit{spatial down-sampling}} and \textbf{\textit{temporal sparse frame sampling}} on the original video. We denote the downsampled aesthetic-specific view of the video as $\mathcal{V} = \{V_i|_{i=0}^{N}\}$, where $V_i$ is the $i$-th frame (in total $N$ frames sampled) of the down-sampled video, with spatial resolution $224\times224$, aligned with the spatial scale during the pre-training of CLIP~\cite{clip}.

\paragraph{Affinity between Video and Texts.} Given any text prompt $T$, the visual ($\mathrm{E}_v$) and textual ($\mathrm{E}_t$) encoders in CLIP extract $\mathcal{V}$ and $T$ into implicit visual ($f_{v,i}$) and textual ($f_t$) features:
\begin{equation}
    f_{v,i} = \mathrm{E}_v(V_i)|_{i = 0}^{N-1}; ~~~~~~
    f_{t}^{T} = \mathrm{E}_t(T)
\end{equation}
Then, the semantic affinity $\mathrm{A}(\mathcal{V}, T)$ between $\mathcal{V}$ (aesthetic view of the video) and text $T$ is defined as follows:
\begin{equation}
    \mathrm{A}(\mathcal{V}, T) = (\sum_{i=0}^{N-1}\frac{f_{v,i}\cdot f_{t}^{T}}{\Vert f_{v,i}\Vert\Vert f_{t}^{T}\Vert})/N
\end{equation}
where the $\cdot$ denotes the dot product of two vectors.

\paragraph{Acronym-Differential Affinity.} In general, a video with good quality should be with higher affinity to \textbf{\green{positive}} quality-related descriptions or feelings ($T_+$, \textit{e.g. ``high quality", ``a good photo", ``clear"}), and lower affinity to \bred{negative} quality-related text descriptions ($T_-$, \textit{e.g. ``low quality", "a bad photo", "unclear"}, acronyms to $T_+$). Therefore, we introduce the Acronym-Differential affinity index ($\mathrm{DA}$), \textit{i.e.} whether the video has higher affinity to positive or negative texts (Fig.~\ref{fig:criterion}(c)), as the semantic criterion for zero-shot VQA:
\begin{equation}
    \mathrm{DA}(\mathcal{V},T_{+},T_{-}) = \mathrm{A}(\mathcal{V}, T_{+}) -  \mathrm{A}(\mathcal{V}, T_{-})
\label{eq:ad}
\end{equation}

\paragraph{Multi-Prompt Aggregation.} As we would like to extract both authentic distortions (which can hardly be detected by NIQE or other low-level indexes) and aesthetic-related issues in the semantic quality index, we aggregate two different pairs of acronyms: \textbf{1)} \textit{high quality}$\leftrightarrow$\textit{low quality}  ($T_{+,0},T_{-,0}$); \textbf{2)} \textit{a good photo}$\leftrightarrow$\textit{a bad photo} ($T_{+,1},T_{-,1}$). Following the advice of VQEG~\cite{vqeg}, we conduct sigmoid remapping to map the two scores into range $[0,1]$ (which is practically similar to human perceptual scales) and sum the remapped scores into the final Semantic Affinity Index ($\mathrm{Q}_{A}$), formalized as follows:
\begin{equation}
    \mathrm{Q}_{A} = \sum_{i=0}^1{\frac{1}{1+e^{-\mathrm{DA}(\mathcal{V},T_{+,i},T_{-,i})}}}
\label{eq:sigmoida}
\end{equation}

\subsection{The Spatial Naturalness Index ($\mathrm{Q}_S$)}
\label{sec:qs} Despite the powerful semantic affinity index, we also utilize the NIQE~\cite{niqe} index, the first completely-blind quality index to detect the traditional types of \textbf{technical distortions}, such as \textit{Additive White Gaussian Noises (AWGN), JPEG compression artifacts}. These distortions are very likely to happen in real-world videos during compression or transmission. To align different indexes, we \textit{normalize} the raw NIQE scores ($\mathrm{Q}_{\mathrm{NIQE}}$ for $V_i$) to Gaussian distribution $N(0,1)$ and rescale them with sigmoid\footnote{As lower raw NIQE/TPQI scores mean better quality, we use negative sigmoid-like remapping $\frac{1}{1+e^x}$ instead of $\frac{1}{1+e^{-x}}$ here (Eq.~\ref{eq:sigmoids}) and in Eq.~\ref{eq:sigmoidt}.} to get the frame-wise naturalness index ($\mathrm{N}_i$):

\begin{equation}
    \mathrm{N}_i = \frac{1}{1 + e^{\frac{\mathrm{Q}_{\mathrm{NIQE},i}-\overline{\mathrm{Q}_{\mathrm{NIQE},i}}}{\sigma(\mathrm{Q}_{\mathrm{NIQE},i})}}}
\label{eq:sigmoids}
\end{equation}
where $\overline{\mathrm{Q}_\mathrm{NIQE}}$ and $\sigma(\mathrm{Q}_\mathrm{NIQE})$ are the \textit{mean} and \textit{standard deviance} of raw NIQE scores in the whole set, respectively. Then, following~\cite{tlvqm,videval,fastervqa}, we sample one frame per second (\textit{1fps}) and calculate the overall \textbf{Spatial Naturalness Index} ($\mathrm{Q}_S$) with sampled frames $V_{F_k}$ in $k$-th second as follows:
\begin{equation}
    \mathrm{Q}_S = \sum_{k=0}^{S_0} \mathrm{N}_{F_k} / S_0
\end{equation}
where $S_0$ is the overall duration of the video. 

\begin{table*}
\footnotesize
\caption{Benchmark evaluation on the proposed BUONA-VISTA, compared with other Opinion-Unaware Quality Indexes.}\label{table:eva}
\label{table:vqc}
\setlength\tabcolsep{6.6pt}
\renewcommand\arraystretch{1.05}
\footnotesize
\centering
\vspace{-8pt}
\resizebox{\textwidth}{!}{\begin{tabular}{l|cc|cc|cc|cc}
\hline
\textbf{Dataset}     & \multicolumn{2}{c|}{LIVE-VQC}   & \multicolumn{2}{c|}{KoNViD-1k}        & \multicolumn{2}{c|}{YouTube-UGC}     &  \multicolumn{2}{c}{CVD2014}           \\ \hline
Methods
        & SRCC$\uparrow$   & PLCC$\uparrow$      & SRCC$\uparrow$   & PLCC$\uparrow$             & SRCC$\uparrow$   & PLCC$\uparrow$                   &SRCC$\uparrow$  & PLCC$\uparrow$                                 \\ \hline 
\rowcolor{lightgray} \multicolumn{9}{l}{\textbf{Opinion-Aware Methods:} }           \\ \hdashline
\rowcolor{lightgray} TLVQM (TIP, 2019) \cite{tlvqm}    & 0.799 &  0.803  & 0.773 & 0.768   & 0.669 &  0.659   & 0.830 &    0.850     \\
\rowcolor{lightgray} VSFA (ACMMM, 2019) \cite{vsfa}         & 0.773 &  0.795  & 0.773 & 0.775   & 0.724 &  0.743   & 0.870 &     0.868 \\
\rowcolor{lightgray} VIDEVAL (TIP, 2021) \cite{videval} &  0.752 &  0.751  & 0.783 & 0.780           & 0.779 &  0.773    & 0.832 &     0.854   \\ 
\hdashline
\multicolumn{9}{l}{\textbf{Existing Opinion-Unaware (\textit{zero-shot}) Approaches:} }           \\\hdashline

(\textit{Spatial}) NIQE (Signal Processing, 2013)~\cite{niqe}  & 0.596 & 0.628 & 0.541 & \blue{0.553} & 0.278 & 0.290 & \blue{0.492} & \blue{0.612} \\
(\textit{Spatial}) IL-NIQE (TIP, 2015)~\cite{ilniqe} & 0.504 & 0.544 & 0.526 & 0.540  & \blue{0.292} & \blue{0.330} & 0.468 & 0.571\\
(\textit{Temporal}) VIIDEO (TIP, 2016)~\cite{viideo} & 0.033 & 0.215 & 0.299 & 0.300  & 0.058 & 0.154 & 0.149 & 0.119 \\
(\textit{Temporal}) TPQI (ACMMM, 2022)~\cite{tpqi}  & \blue{0.636} & \blue{0.645} & \blue{0.556} & 0.549 & 0.111 & 0.218 & 0.408 & 0.469 \\ 
\hdashline

\rowcolor{lightpink} \textbf{BUONA-VISTA (Ours, \textit{zero-shot})} & \bred{0.784} & \bred{0.794} & \bred{0.760} & \bred{0.760} & \bred{0.525} & \bred{0.556} & \bred{0.740} & \bred{0.763}\\ 
\rowcolor{lightpink} \textit{Improvements to Existing Best} & 23\% & 23\% & 37\% & 38\% & 80\% & 69\% & 50\% & 25\% \\
\hline
\end{tabular}}
\vspace{-13pt}
\end{table*}

\subsection{The Temporal Naturalness Index ($\mathrm{Q}_T$)}
\label{sec:qt}

While the $\mathrm{Q}_A$ and $\mathrm{Q}_S$ can better cover different types of spatial quality issues, they are unable to cover the distortions in the temporal dimension, such as \textit{shaking}, \textit{stall}, or \textit{unsmooth camera movements}, which are well-recognized~\cite{cnn+lstm,tlvqm,deepvqa,discovqa} to affect the human quality perception. In general, all these temporal distortions can be summarized as non-smooth inter-frame changes between adjacent frames, and can be captured via recently-proposed TPQI~\cite{tpqi}, which is based on the neural-domain curvature across three continuous frames. Specifically, the curvatures can be computed via the simulated neural responses on the primary visual cortex (V1,~\cite{primaryv1}) and lateral geniculate nucleus (LGN,~\cite{lgn}) domains, as follows:
\begin{equation}
    \mathrm{\mathrm{Q}_{TPQI}} = \frac{\log{(\frac{1}{M-2}\sum_{j=1}^{M-2})\mathbb{C}^\mathrm{V1}_j} + \log{(\frac{1}{M-2}\sum_{j=1}^{M-2})\mathbb{C}^\mathrm{LGN}_j}}{2}
\end{equation}
where $M$ is the total number of frames in the whole video, $\mathbb{C}_\mathrm{LGN}$ and $\mathbb{C}_\mathrm{V1}$ are the curvatures at frame $j$ respectively. The \textbf{Temporal Naturalness Index} ($\mathrm{Q}_T$) is then mapped from the raw scores via gaussian normalization and sigmoid rescaling:

\begin{equation}
    \mathrm{Q}_T = \frac{1}{1 + e^{\frac{\mathrm{Q}_{\mathrm{TPQI}}-\overline{\mathrm{Q}_{\mathrm{TPQI}}}}{\sigma(\mathrm{Q}_{\mathrm{TPQI}})}}}
\label{eq:sigmoidt}
\end{equation}

\begin{table*}[]
\setlength\tabcolsep{4.4pt}
\renewcommand\arraystretch{1.16}
\caption{Comparing the cross-dataset performances of existing opinion-aware approaches with \textbf{\textit{zero-shot}} BUONA-VISTA (which requires no training at all). BUONA-VISTA is notably more robust than these approaches.} \label{tab:crossvsbv}
\vspace{-17pt}
\center
\footnotesize
\resizebox{\linewidth}{!}{\begin{tabular}{l|cc|cc|cc|cc|cc|cc}
\hline
\textbf{Train} on (\textit{None} for BUONA-VISTA) & \multicolumn{4}{c|}{{KoNViD-1k}} & \multicolumn{4}{c|}{{LIVE-VQC}} & \multicolumn{4}{c}{{Youtube-UGC}} \\ \hline
\textbf{Test} on & \multicolumn{2}{c|}{{LIVE-VQC}} & \multicolumn{2}{c|}{{Youtube-UGC}}  & \multicolumn{2}{c|}{{KoNViD-1k}} & \multicolumn{2}{c|}{{Youtube-UGC}}   & \multicolumn{2}{c|}{{LIVE-VQC}} & \multicolumn{2}{c}{{KoNViD-1k}} \\ \hline
\textit{}                         & SRCC$\uparrow$                        & PLCC$\uparrow$                        & SRCC$\uparrow$                        & PLCC$\uparrow$                 & SRCC$\uparrow$                        & PLCC$\uparrow$                        & SRCC$\uparrow$                        & PLCC$\uparrow$        & SRCC$\uparrow$                        & PLCC$\uparrow$               & SRCC$\uparrow$                       & PLCC$\uparrow$                                         \\ 
\hline
CNN-TLVQM (2020,MM)\cite{cnntlvqm}             & 0.713                      & 0.752    & \gray{NA} &   \gray{NA}               & 0.642 & 0.631 & \gray{NA} & \gray{NA}       & \gray{NA}                        & \gray{NA}             & \gray{NA}  &   \gray{NA}                                 \\ 
GST-VQA (2021, TCSVT)\cite{gstvqa}             & 0.700                      & 0.733    & \gray{NA} &   \gray{NA}               & \blue{0.709} & 0.707 & \gray{NA} & \gray{NA}       & \gray{NA}                        & \gray{NA}             & \gray{NA}  &   \gray{NA}                                 \\ 
VIDEVAL (2021, TIP)\cite{videval}                    & 0.627                         & 0.654   & 0.370 & 0.390                  & 0.625                        & 0.621 & 0.302 & 0.318  & 0.542                         & 0.553      & 0.610 &   0.620                           \\ 
MDTVSFA (2021, IJCV)\cite{mdtvsfa}                     & \blue{0.716}                        & \blue{0.759}    & \blue{0.408} &  \blue{0.443}                 & 0.706                         & \blue{0.711}  & \blue{0.355} &  \blue{0.388}      & \blue{0.582}                        & \blue{0.603}   & 0.649 &   0.646                          \\ \hdashline
\rowcolor{lightpink} \textbf{BUONA-VISTA (\textit{zero-shot})} & \bred{0.784}&\bred{0.794}&\bred{0.525}&\bred{0.556}&\bred{0.760}&\bred{0.760}&\bred{0.525}&\bred{0.556}&\bred{0.784}&\bred{0.794}&\bred{0.760}&\bred{0.760}\\ \hline
\end{tabular}}
\vspace{-7pt}
\end{table*}

\subsection{\textit{BUONA-VISTA} Index: Metric Aggregation}
As we aim to design a robust opinion-unaware perceptual quality index, we directly aggregate all the indexes by summing up the scale-aligned scores without regression from any VQA datasets. As the $\mathrm{Q}_A$, $\mathrm{Q}_S$ and $\mathrm{Q}_T$ have already been gaussian-normalized and sigmoid-rescaled in Eq.~\ref{eq:sigmoida}, Eq.~\ref{eq:sigmoids} and Eq.~\ref{eq:sigmoidt} respectively, all three metrics are in range $[0,1]$, and the overall unified \textbf{BUONA-VISTA} index $\mathrm{Q}_\text{Unified}$ is defined as:

\begin{equation}
    \mathrm{Q}_\text{Unified} = \mathrm{Q}_A + \mathrm{Q}_S + \mathrm{Q}_T
\end{equation}

In the next section, we will conduct several experimental studies to prove the effectiveness of each separate index and the rationality of the proposed aggregation strategy.

\section{Experimental Evaluations}

\subsection{Implementation Details}

Due to the differences of the targeted quality-related issues in three indexes, the inputs of the three branches are different. For $\mathrm{Q}_A$, the video is spatially downsampled to $224\times 224$ via a bicubic~\cite{bicubic} downsampling kernel, and temporally sub-sampled to $N=32$ uniform frames~\cite{dover}. For $\mathrm{Q}_S$, the video retains its original spatial resolution but temporally only keep $S_0$ uniform frames, where $S_0$ is the duration of the video (\textit{unit: second}). For $\mathrm{Q}_T$, all videos are spatially downsampled to $270\times480$ (to keep the aspect ratio), with all frames fed into the neural response simulator.  The $\mathrm{Q}_A$ is calculated with Python 3.8, Pytorch 1.7, with official CLIP-ResNet-50~\cite{he2016residual} weights. The $\mathrm{Q}_S$ and $\mathrm{Q}_T$ are calculated with Matlab R2022b. 

\subsection{Evaluation Settings}

\paragraph{Evaluation Metrics.} Following common studies, we use two metrics, the Spearman Rank-order Correlation Coefficients (SRCC) to evaluate monotonicity between quality scores and human opinions, and the Pearson Linearity Correlation Coefficients (PLCC) to evaluate linear accuracy. 

\paragraph{Benchmark Datasets.} To better evaluate the performance of the proposed BUONA-VISTA under different in-the-wild settings, we choose four different datasets, including CVD2014~\cite{cvd} (234 videos, with lab-collected authentic distortions during capturing), LIVE-VQC~\cite{vqc} (585 videos, recorded by smartphones), KoNViD-1k~\cite{kv1k} (1200 videos, collected from social media platforms), and YouTube-UGC~\cite{ytugc,ytugccc} (1131 available videos, non-natural videos collected from YouTube with categories \textit{Gaming/Animation/Lyric Videos}).

\begin{table*}[]
\caption{Ablation Studies (I): effects of different indexes in the proposed BUONA-VISTA, on three natural video datasets.}
\vspace{-8pt}
\renewcommand\arraystretch{1.10} 
\resizebox{\textwidth}{!}{
\begin{tabular}{ccc|cc|cc|cc}
\hline
 \multicolumn{3}{c|}{Different Indexes in BUONA-VISTA}                    & \multicolumn{2}{c|}{LIVE-VQC} & \multicolumn{2}{c|}{KoNViD-1k} & \multicolumn{2}{c}{CVD2014} \\ \hline
Semantic Affinity ($\mathrm{Q}_A$) & Spatial Naturalness ($\mathrm{Q}_S$) & Temporal Naturalness ($\mathrm{Q}_T$) & SRCC$\uparrow$       & PLCC$\uparrow$   & SRCC$\uparrow$       & PLCC$\uparrow$   & SRCC$\uparrow$       & PLCC$\uparrow$         \\
\hline
\cmark &   &   & 0.629 & 0.638 & 0.608 & 0.602 & 0.685 & 0.692  \\
  & \cmark &   & 0.593 & 0.615 & 0.537 & 0.528 & 0.489 & 0.558 \\
   &   & \cmark & 0.690 & 0.682& 0.577 & 0.569&0.482&0.498 \\
\cdashline{1-9}
 \cmark & \cmark & & 0.692 & 0.712 & 0.718 & 0.713 & 0.716 & 0.731 \\
   & \cmark & \cmark & 0.749 & 0.753 & 0.670 & 0.672 & 0.618 & 0.653 \\
 \cmark &   & \cmark & 0.767  & 0.768 & 0.704 & 0.699 & 0.708 & 0.725\\
\cdashline{1-9}
\rowcolor{lightpink} \cmark & \cmark & \cmark & \bred{0.784} & \bred{0.794} & \bred{0.760} & \bred{0.760} & \bred{0.740} & \bred{0.763} \\
\hline
\end{tabular}}
\label{tab:ablation}
\vspace{-4mm}
\end{table*}
\begin{table*}[]
\caption{Ablation Studies (IV): effects of different text prompts and the proposed multi-prompt aggregation strategy.}
\vspace{-8pt}
\renewcommand\arraystretch{1.22} 
\setlength\tabcolsep{6pt}
\resizebox{\linewidth}{!}{
\begin{tabular}{c|c|c|c|c|c|c|c}
\hline
& \multicolumn{3}{c|}{Overall Performance of BUONA-VISTA} & \multicolumn{4}{c}{Performance of Semantic Affinity Index Only} \\
\hline
{Dataset}                    & {LIVE-VQC} & {KoNViD-1k} & {CVD2014} & {LIVE-VQC} & {KoNViD-1k} & {CVD2014} & {YouTube-UGC} \\ \hline
\textbf{Prompt Pairs} & SRCC$\uparrow$/PLCC$\uparrow$   & SRCC$\uparrow$/PLCC$\uparrow$   & SRCC$\uparrow$/PLCC$\uparrow$  &SRCC$\uparrow$/PLCC$\uparrow$   & SRCC$\uparrow$/PLCC$\uparrow$   & SRCC$\uparrow$/PLCC$\uparrow$     & SRCC$\uparrow$/PLCC$\uparrow$           \\ \hline
\textbf{(a)} \textit{[high $\leftrightarrow$low] quality}   & 0.768/0.775 & 0.725/0.725 & 0.738/0.757 &0.560/0.575&0.477/0.472&\bred{0.728}/\bred{0.729}&0.539/0.564\\
\textbf{(b)} \textit{a [good$\leftrightarrow$bad] photo} & 0.778/0.785 & 0.727/0.727 & 0.653/0.686 &0.608/0.581&0.586/0.551&0.507/0.512&0.473/0.458 \\ 
\hdashline
\rowcolor{lightpink} \textbf{(a)}+\textbf{(b)} \textit{Aggregated} & \bred{0.784}/\bred{0.794} & \bred{0.760}/\bred{0.760} & \bred{0.740}/\bred{0.763} &\bred{0.629}/\bred{0.638}&\bred{0.609}/\bred{0.602}&0.686/0.693&\bred{0.585}/\bred{0.606} \\
\hline
\end{tabular}}
\label{table:prompt}
\vspace{-14pt}
\end{table*}

\subsection{Benchmark Comparison}

We compare with both existing opinion-unaware (zero-shot) or opinion-aware VQA methods in Tab.~\ref{table:eva}/\ref{tab:crossvsbv} to evaluate the accuracy and robustness of proposed BUONA-VISTA index.

\paragraph{Comparison with Opinion-Unaware Approaches.} The proposed BUONA-VISTA quality index is notably better than any existing opinion-unaware quality indexes with \textbf{\textit{at least 20\%}} improvements. Specifically, on the three natural VQA datasets (LIVE-VQC, KoNViD-1k and CVD2014), it has reached almost 0.8 PLCC/SRCC, which are even on par with or better than some opinion-aware approaches. On the non-natural dataset (YouTube-UGC), with the assistance of the semantic affinity index, the proposed BUONA-VISTA has extraordinary \textbf{80\%} improvement than all semantic-unaware zero-shot quality indexes, \textit{for the first time} provides reasonable quality predictions on this dataset. Without any training, these results demonstrate that the proposed BUONA-VISTA achieves leapfrog improvements over existing metrics and can be widely applied as a robust real-world video quality metric.

\begin{figure}
    \centering
\includegraphics[width=\linewidth]{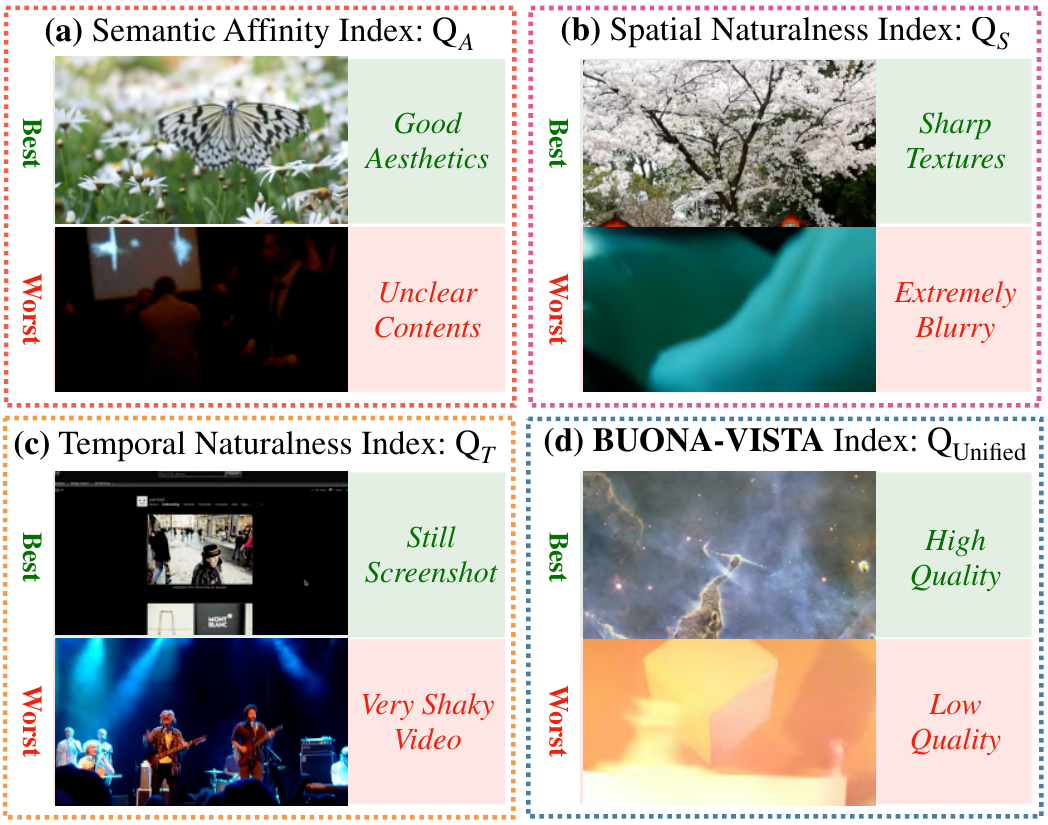}
    \vspace{-18pt}
    \caption{Videos with \textit{best/worst} quality in perspective of three separate indexes, and the overall BUONA-VISTA index. All demo videos are appended in \textbf{\textit{supplementary materials}}.}
    \label{fig:vis}
    \vspace{-18pt}
\end{figure}

\paragraph{Comparison with Opinion-Aware Approaches.} Though it is extremely difficult, if not impossible for BUONA-VISTA to surpass opinion-aware approaches, it has largely bridged the gap between zero-shot and supervised methods. Moreover, these opinion-aware methods might face an extra challenge of over-fitting on specific datasets. As compared between their Cross-dataset Performances and results of BUONA-VISTA in Tab.~\ref{tab:crossvsbv} the proposed \textbf{\textit{zero-shot}} BUONA-VISTA can out-perform existing opinion-based methods when they do not train and test on the same set of videos and opinions, further proving the robustness of the proposed BUONA-VISTA.

\begin{table}[]
\caption{Ablation Studies (II): effects of different indexes in the proposed BUONA-VISTA on YouTube-UGC dataset.}
\vspace{-8pt}
\centering
\renewcommand\arraystretch{1.0} 
\setlength\tabcolsep{14pt}
\resizebox{0.9\linewidth}{!}{
\begin{tabular}{ccc|cc}
\hline
 \multicolumn{3}{c|}{Indexes in BUONA-VISTA}                 & \multicolumn{2}{c}{YouTube-UGC} \\ \hline
 $\mathrm{Q}_A$ &   $\mathrm{Q}_S$ & $\mathrm{Q}_T$ & SRCC$\uparrow$       & PLCC$\uparrow$ \\
\hline
\cmark &   &   & 0.585 & \textbf{0.606} \\
\cmark & \cmark  &   & \textbf{0.589} & 0.604 \\
\rowcolor{lightpink} \cmark & \cmark  & \cmark  & 0.525 & 0.556 \\ \hdashline
& \cmark  &   & 0.240 & 0.153 \\
& & \cmark & 0.133 & 0.141 \\
\hline
\end{tabular}}
\vspace{-10pt}
\label{table:ugc}
\end{table}

\begin{table}[]
\caption{Ablation Studies (III): comparison of different aggregation strategies in the proposed BUONA-VISTA.}
\vspace{-8pt}
\renewcommand\arraystretch{1.22} 
\setlength\tabcolsep{3pt}
\resizebox{\linewidth}{!}{
\begin{tabular}{c|c|c|c}
\hline
{\textbf{Aggregation}}                    & {LIVE-VQC} & {KoNViD-1k} & {CVD2014} \\ \hline
Metric & SRCC$\uparrow$/PLCC$\uparrow$   & SRCC$\uparrow$/PLCC$\uparrow$   & SRCC$\uparrow$/PLCC$\uparrow$         \\ \hline
\textit{Direct Addition} & 0.760/0.750 & 0.675/0.660 & 0.664/0.699 \\
\textit{Linear} + \textit{Addition} & 0.776/0.760 & 0.720/0.710  & 0.700/0.729 \\ \hdashline
 \textit{Sigmoid} + \textit{Multiplication} &  0.773/0.729 & 0.710/0.679 & 0.692/0.661 \\
\hdashline
\rowcolor{lightpink} \textit{Sigmoid} + \textit{Addition} & \bred{0.784}/\bred{0.794} & \bred{0.760}/\bred{0.760} & \bred{0.740}/\bred{0.763} \\
\hline
\end{tabular}}
\label{table:agg}
\vspace{-10pt}
\end{table}

\subsection{Qualitative Studies}

In the qualitative studies, we visualize snapshots of videos with highest or lowest score in each separate index, and the overall BUONA-VISTA index. As shown in Fig.~\ref{fig:vis}, the \textbf{(a)} Semantic Affinity is highly related to \textbf{\textit{aesthetics}}, where the \textbf{(b)} Spatial Naturalness focus on spatial textures (\textit{sharp}$\leftrightarrow$\textit{blurry}), and the \textbf{(c)} Temporal Naturalness focus on temporal variations (\textit{stable}$\leftrightarrow$\textit{shaky}), aligning with the aforementioned criteria of the three indexes. We also append the original videos of these examples in our \textbf{\textit{supplementary materials}}.

\subsection{Ablation Studies}

In the ablation studies, we discuss the effects of different quality indexes: Semantic Affinity, Spatial Naturalness and Temporal Naturalness (Sec.~\ref{sec:ablsi}), on either natural. We then discuss the effects of the aggregation strategies (Sec.~\ref{sec:ablas}). Moreover, we evaluate the effects of different prompt pairs and the proposed multi-prompt aggregation (Sec.~\ref{sec:ablprompt}).

\subsubsection{Effects of Separate Indexes}
\label{sec:ablsi}

\paragraph{Evaluation on Natural Datasets.} During evaluation on the effects of separate indexes, we divide the four datasets into two parts: for the first part, we categorize the LIVE-VQC, KoNViD-1k and CVD2014 as \textbf{natural datasets}, as they do not contain computer-generated contents, or movie-like edited and stitched videos. We list the results of different settings in Tab.~\ref{tab:ablation}, where all three indexes contribute notably to the final accuracy of the proposed BUONA-VISTA, proving that the semantic-related quality issues, traditional spatial distortions and temporal distortions are all important to building an robust estimation on human quality perception. Specifically, in CVD2014, where videos only have authentic distortions during capturing, the Semantic Affinity ($\mathrm{Q}_A$) index shows has largest contribution; in LIVE-VQC, the dataset commonly-agreed with most temporal distortions, the Temporal Naturalness ($\mathrm{Q}_T$) index contributes most to the overall accuracy. These results demonstrate our aforementioned claims on the separate concerns of the three indexes.

\paragraph{Evaluation on YouTube-UGC.} In YouTube-UGC, as shown in Tab.~\ref{table:ugc}, the Spatial Naturalness index cannot improve the final performance of the BUONA-VISTA, where the Temporal Naturalness index even lead to 8\% performance drop. As YouTube-UGC are all long-duration (20-second) videos and almost every videos is made up of multiple scenes, we suspect this performance degradation might come from the during scene transition, where the temporal curvature is very large but do not lead to degraded quality. In our future works, we consider detecting scene transition in videos and only compute the Temporal Naturalness Index within the same scene. 

\subsubsection{Effects of Aggregation Strategies}
\label{sec:ablas}

We evaluate the effects of aggregation strategies in Tab.~\ref{table:agg}, by comparing with different rescaling strategies (\textit{Linear} denotes Gaussian Noramlization only, and \textit{Sigmoid} denotes Gaussian followed by Sigmoid Rescaling) and different fusion strategies (\textit{addition($+$) or multiplication($\times$)}). The results have demonstrated that the both gaussian normalization and sigmoid rescaling contributes to the final performance of aggregated index, and \textit{addition} is better than \textit{multiplication}.
\subsubsection{Effects of Different Text Acronym Pairs}
\label{sec:ablprompt}

In Tab.~\ref{table:prompt}, we discuss the effects of different text acronym pairs as $T_{+}$ and $T_{-}$ in Eq.~\ref{eq:ad}. We notice that \textit{[high$\leftrightarrow$low] quality} can achieve very good performance on CVD2014, where the content diversity can be neglected and the major concern is the \textbf{authentic distortions}. For LIVE-VQC and KoNViD-1k (with diverse aesthetics), however, the \textit{[good$\leftrightarrow$bad] photo} prompt shows higher accuracy. The results suggests that different datasets have different quality concerns, while aggregating two acronym pairs can result in stable improvements for overall performance in all datasets, proving the effectiveness of the proposed multi-prompt aggregation strategy.

\section{Conclusion}

In this paper, we propose BUONA-VISTA, a robust zero-shot opinion-unaware video quality index for in-the-wild videos, which aligned and aggregated CLIP-based text-prompted semantic affinity index with traditional technical metrics on spatial and temporal dimensions. The proposed BUONA-VISTA achieves unprecedented performance among opinion-unaware video quality indexes, and demonstrates better robustness than opinion-aware VQA approaches across different datasets. We hope the proposed robust video quality index can serve as an reliable and effective metric in related researches on videos and contribute in real-world applications.

{\small
\bibliographystyle{IEEEtran}
\bibliography{egbib}
}

\end{document}